\documentclass[final,runningheads]{llncs}
\usepackage{graphicx}
\usepackage{hyperref}

\usepackage{amsmath,amssymb,amsfonts}
\usepackage{siunitx,csquotes}
\usepackage{url}
\usepackage{booktabs}
\usepackage[dvipsnames]{xcolor}
\definecolor{mint}{rgb}{0.24, 0.71, 0.54}
\definecolor{midnightblue}{rgb}{0.1, 0.1, 0.44}
\definecolor{moonstone}{rgb}{0.45, 0.66, 0.76}

\usepackage{lastpage} 
\usepackage[llncs]{llncsconf}
\llncsdoi{10.1007/978-3-030-61616-8_49}
\llncs{Published in \emph{Artificial Neural Networks and Machine Learning -- ICANN 2020, Part II}. Springer International Publishing}{610}

\begin{document}
\title{Benchmarking Deep Spiking Neural Networks on Neuromorphic Hardware}%
\author{Christoph Ostrau\orcidID{0000-0003-1551-3848}\inst{1}\and
Jonas Homburg\orcidID{0000-0002-4760-8510}\inst{1} \and
Christian Klarhorst\inst{1} \and
Michael Thies\inst{1} \and Ulrich Rückert\inst{1}}
\authorrunning{C. Ostrau et al.}
\institute{Technical Faculty, Bielefeld University, Bielefeld, Germany
\email{costrau@techfak.uni-bielefeld.de}}
\maketitle              %
\begin{abstract}
With more and more event-based neuromorphic hardware systems being developed at universities and in industry, there is a growing need for assessing their performance with domain specific measures.
In this work, we use the methodology of converting pre-trained non-spiking to spiking neural networks to evaluate the performance loss and measure the energy-per-inference for three neuromorphic hardware systems (BrainScaleS, Spikey, SpiNNaker) and common simulation frameworks for CPU (NEST) and CPU/GPU (GeNN).
For analog hardware we further apply a re-training technique known as hardware-in-the-loop training to cope with device mismatch. %
This analysis is performed for five different networks, including  three networks that have been found by an automated optimization with a neural architecture search framework.
We demonstrate that the conversion loss is usually below one percent for digital implementations, and moderately higher for analog systems with the benefit of much lower energy-per-inference costs. 

\keywords{Spiking Neural Networks  \and Neural Architecture Search \and Benchmark.}
\end{abstract}

\section{Introduction}

Diverse event-based neuromorphic hardware systems promise the accelerated execution of so called spiking neural networks (SNN), also referred to as the third generation of neural networks \cite{Maass1997}. 
The most prominent representatives of this class of hardware accelerators include the platforms Braindrop \cite{Neckar2019}, BrainScaleS \cite{schemmel2010waferscale}, DYNAPs \cite{moradi2018scalable}, Loihi \cite{davies2018loihi}, SpiNNaker \cite{Furber2013} and Truenorth \cite{Akopyan2015}.
With the diversity of hardware accelerators comes a problem for potential end-users: 
which platform is suited best for a given spiking neural network algorithm, possibly respecting inherent resource requirements for embedding in mobile robots or smart devices.
Usually, this question is answered by evaluating a set of benchmarks on all qualified systems, which measure the state-of-the-art and quantify progress in future hardware generations (see e.g. \cite{davies2019benchmarks})).
Here, we face two major challenges with neuromorphic hardware.
First, there is no universal interface to all hardware/software simulators despite some projects like PyNN \cite{Davison2008}.
Second, there are quite a few promising network models and learning strategies, but still \enquote{the} algorithm for spiking neural networks is missing.
One recent system overarching network is the cortical microcircuit model \cite{VanAlbada2018,Knight2018}.
A follow-up publication \cite{Rhodes2020} shows, how this benchmark has driven platform specific optimization that, in the end, improves the execution of various networks on the SpiNNaker platform confirming the value of benchmarks.
However, it is also an example of a platform specific implementation to reach maximal performance on a given system.

One commonly agreed application for spiking neural networks is the conversion of conventionally trained artificial neural networks (ANN) to rate-based SNNs \cite{Diehl2015}.
Although this is not using SNNs in their most efficient way, it is a pragmatic approach that is suitable to be ported to different accelerators, independent of their nature.
In this work, we use this approach for evaluating five distinct networks, either defined by hardware restrictions, by already published work, or by employing neural architecture search (NAS) with Lamarck\_ML \cite{Homburg2019} to optimize the network topology.
We evaluate these networks on BrainScaleS, Spikey \cite{pfeil2013six}, and SpiNNaker as well as the CPU simulator NEST \cite{gewaltig2007nest} and the CPU/GPU code-generation framework GeNN \cite{Yavuz2016}.
Furthermore, we use a retraining approach with neuromorphic hardware-in-the-loop (HIL) proposed in \cite{schmitt2017neuromorphic} to unlock the full potential of the analog neuromorphic hardware systems.
Section 2 outlines the target systems, the software environment, and the used methods.
Section 3 presents the results, including neuron parameter optimization, and accuracy along with energy measurements for all target platforms.

\section{Methods}
In the following we introduce all target systems and the software environment as well as the methodology followed.

\subsection{Target Systems and Software}
    All target systems in this work support the simulation or emulation of leaky integrate-and-fire neurons with conductance-based synapses, although especially analog systems are limited to specific neuron models.
    \textbf{NEST} is a scaleable software simulator suited to simulate small as well as extensive networks on compute clusters. %
    It is used in version 2.18 \cite{Jordan2019} executed with four threads on an Intel Core i7-4710MQ mobile processor.
    \textbf{GeNN} \cite{Yavuz2016} is a code generation framework for the simulation of SNNs.
    In its current release version (4.2.1)\footnote{Here, we use the most recent GeNN from github (end of April 2020)}, it supports generating code for a single-threaded CPU simulation or for graphics processing units (GPU) supporting NVIDIA CUDA.
    Networks are evaluated on a NVIDIA GeForce 1080 TI GPU; runtimes are measured for networks without recording any spikes due to the overhead of getting spikes back from GPU, which effectively stops the simulation at every time step and copies the data between GPU and CPU. 
    For this publication we make use of single precision accuracy and all simulators use a time step of \SI{1}{\milli\second}. 
    However, NEST is using an adaptive time-step to integrate the neuron model.
    The fully digital many-core architecture \textbf{SpiNNaker} \cite{Furber2013} comes in two different sizes, which are both used in this work.
    The smaller SpiNN3 system is composed of four chips; the larger SpiNN5 board consists of 48 chips.
    A single chip comprises 18 ARM968 general purpose CPU cores, with each simulating up to 255 \verb|IF_cond_exp| neurons.
    The system runs in real-time, simulating \SI{1}{\milli\second} of model time in \SI{1}{\milli\second} wall clock time.
    SpiNNaker is used with the latest released software version 5.1.0 using PyNN 0.9.4.
    Finally, we make use of two mixed-signal (analog neural circuits, digital interconnect) systems:
    First, the \textbf{Spikey} system \cite{pfeil2013six} supports the emulation of 384 neurons with 256 synapses each. 
    The emulated neuron model is subject to restricted parameter ranges (e.g. four bit weights, limited time constants) with some parameters prescribed by the hardware (e.g. the membrane capacitance).
    The system runs at a speedup of $10,000$, therefore taking only \SI{0.1}{\micro\second} to emulate \SI{1}{\milli\second} of model time.
    Second, Spikey's successor \textbf{BrainScaleS} \cite{schemmel2010waferscale} shares many of Spikey's properties. 
    Most notably is the now fully parameterizable neuron model, as well as the usage of wafer-scale integration, combining 384 accessible HICANN chips on a single wafer for a full system.
    Each chip implements 512 neuron circuits with 220 synapses each, where up to 64 circuits can be combined to form a single virtual neuron, allowing more robust emulations and a higher synapse fan-in.
    
    While all of these platforms formally support the \textbf{PyNN} API \cite{Davison2008}, the supported API versions differ between simulators impeding the portability of code. 
    \textbf{Cypress}\footnote{\url{https://github.com/hbp-unibi/cypress}} \cite{stockel2017binary} is a C++ framework abstracting away these differences. 
    For NEST, Spikey and SpiNNaker the framework makes use of their PyNN interfaces, however, for BrainScaleS and GeNN a lower-level C++ interface is used.
    Furthermore, the proposed networks studied below are part of the \textbf{S}piking \textbf{N}eural \textbf{A}rchitecture \textbf{B}enchmark \textbf{Suite}\footnote{The code for this and other work can be found at \url{https://github.com/hbp-unibi/SNABSuite}} (SNABSuite)\cite{2941207,2941831}, which also covers benchmarks like low-level synthetic characterizations and application-inspired (sub-)tasks with an associated framework for automated evaluation.
    
    Energy measurements have been taken with a Ruideng UM25C power meter (SpiNNaker, Spikey), with a PeakTech 9035 for CPU simulations, or with the NVIDIA \texttt{smi} tool.
    There is no possibility for remote energy measurements on the BrainScaleS system. Thus, the values have been estimated from the number of pre-synaptic events using published data in \cite{schmitt2017neuromorphic}.

\subsection{Converting DNNs to SNNs}
\begin{figure}[t]
        \centering
        \includegraphics[]{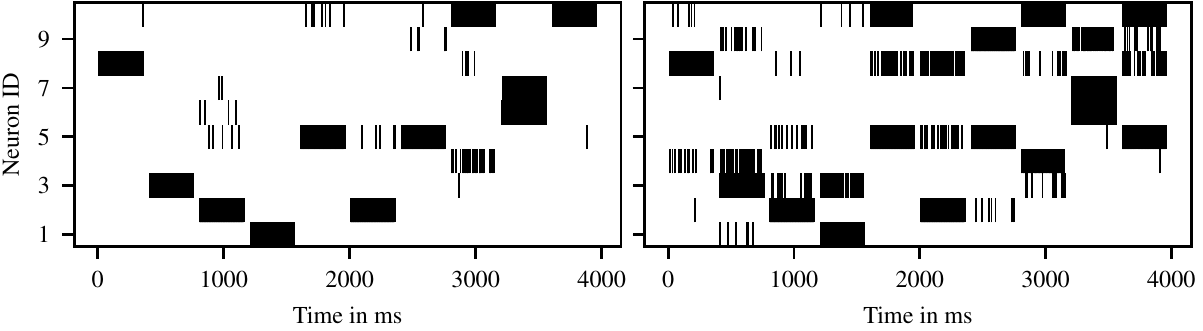}
        \caption{Output spikes for converted networks. Left: Output spikes of a network that has been trained using a softmax layer as the last layer. Right: The same network trained with only ReLU activation functions.}
        \label{fig:reluvssoft}
    \end{figure}

    This work is based on the idea of \cite{Cao2015,Diehl2015}, where a pre-trained artificial neural network is converted into a SNN.
    In this case, we train several multi-layer perceptrons that differ in size to classify MNIST handwritten digits.
    The training uses standard batch-wise gradient-descent in combination with error back-propagation.
    The conversion method exploits that the activation curve of a LIF neuron resembles the ReLU activation curve, such that float (analog) values of the ANN become spike rates in the SNN.
    All weights of the ANN are normalized to the maximal weight of the full network, and then scaled to a maximal value either given by restrictions of the hardware platform (e.g. 4 bit weights on Spikey/BrainScaleS) or  determined by parameter optimization (see below for details).
    Similarly, other parameters of the SNN are found by extensive parameter tuning or are fixed due to hardware constraints.
    Neuron biases are not easily and efficiently mapped to SNNs, which is why we set all bias terms to zero in the training process of the ANN. 
    In contrast to \cite{Diehl2015}, we found that using a soft-max layer as the last layer in the ANN for training does not necessarily decrease the performance of the SNN. 
    However, using soft-max will lead to an increased number of spikes for all rejected classes (cf. Figure \ref{fig:reluvssoft}).
    {}
    
    As the Spikey platform is very limited in size and connectivity, the smallest and simplest network (referred to as \textit{Spikey network}) consists of a single hidden layer with 100 neurons and no inhibitory connections. Spikey requires separation of excitation and inhibition  at the neuron level and consists of two separate chips with limited connectivity between them.
    Thus, we only used positive weights and achieved the best performance using a hinge loss, which increases the weights for the winner neurons and decreases weights for the second place neuron only.
    Due to the acceleration factor of Spikey and BrainScaleS, communication bandwidth limits the usable spike rates.
    Too high rates (input as well as inter-neuron rates) will inevitably lead to spike loss that would reduce the performance of the network.
    This naturally restricts the parameter space to be evaluated.
    Still, there is a significant performance loss when applying the conversion process for analog systems.
    Perfect conversion requires that every synapse with the same weight and every neuron behaves in the same way, referring to identical activation curves.
    On analog systems, however, we have to deal with temporal noise perturbing the membrane voltage, trial-to-trial variation and analog mismatch between circuits \cite{Petrovici2014a}.
    As shown in \cite{stockel2017binary}, such a hardware network will perform at roughly 60-70\% accuracy compared to a simulator, even after platform specific parameter tuning.
    \cite{schmitt2017neuromorphic} proposed to train the pre-trained neural network again while replacing the outputs of the ANN with spike rates recorded from hardware employing back-propagation to train a device specific network.
    All details can be found in \cite{schmitt2017neuromorphic} (Figure 7).
    
\subsection{Neural Architecture Search (NAS)}
    Lamarck\_ML\footnote{\url{https://github.com/JonasDHomburg/LAMARCK_ML}}\cite{Homburg2019} is a modular and extensible Python library for application driven exploration of network architectures.
    This library allows to define a class of network architectures to be examined and operations to modify and combine those architectures.
    These definitions are then used by a search algorithm to explore and evaluate network architectures in order to maximize an objective function.
    For this work, the limitations of the neuromorphic hardware systems compared to state-of-the-art processing units are the leading motivation for the applied restrictions. %
    The applied layer types are limited to fully connected layers which may be arranged in a nonsequential manner resulting in an acyclic directed graph structure.
    To preserve the structural information of a single neural network in the exploration process, a meta graph is created to contain the current network and the meta graph of the networks which were involved in creating it.
    This process is unbounded and accumulates structural information over several generations in the meta graph.
    To forget unprofitable information, the meta graph is designed to dismiss structural information that has not been used in the last five exploration steps.
    One exploration step consists of combining the meta graph of two network architectures and sampling a new path in this meta graph in order to create an improved architecture.
    A new architecture is created by sampling a path based on the quality of its best architecture and amending it with elements that have not been examined before.
    
    The exploration procedure is performed by a genetic algorithm configured with a generation size of 36 network architectures of which 20 are selected based on an exponential ranking to create new architectures for the next generation.
    This next generation is created with an elitism replacement scheme that preserves the best two network architectures of the previous generation.
    In total 75 generations have been created in the NAS to find an architecture that achieves at least 97\% evaluation accuracy. 
    Above this threshold, an architecture is defined to be better if it requires less than 100 neurons for increasing the accuracy by 1\%. \looseness=-1
\section{Results}
    \begin{figure}[t]
        \centering
        \includegraphics[width=0.48\textwidth]{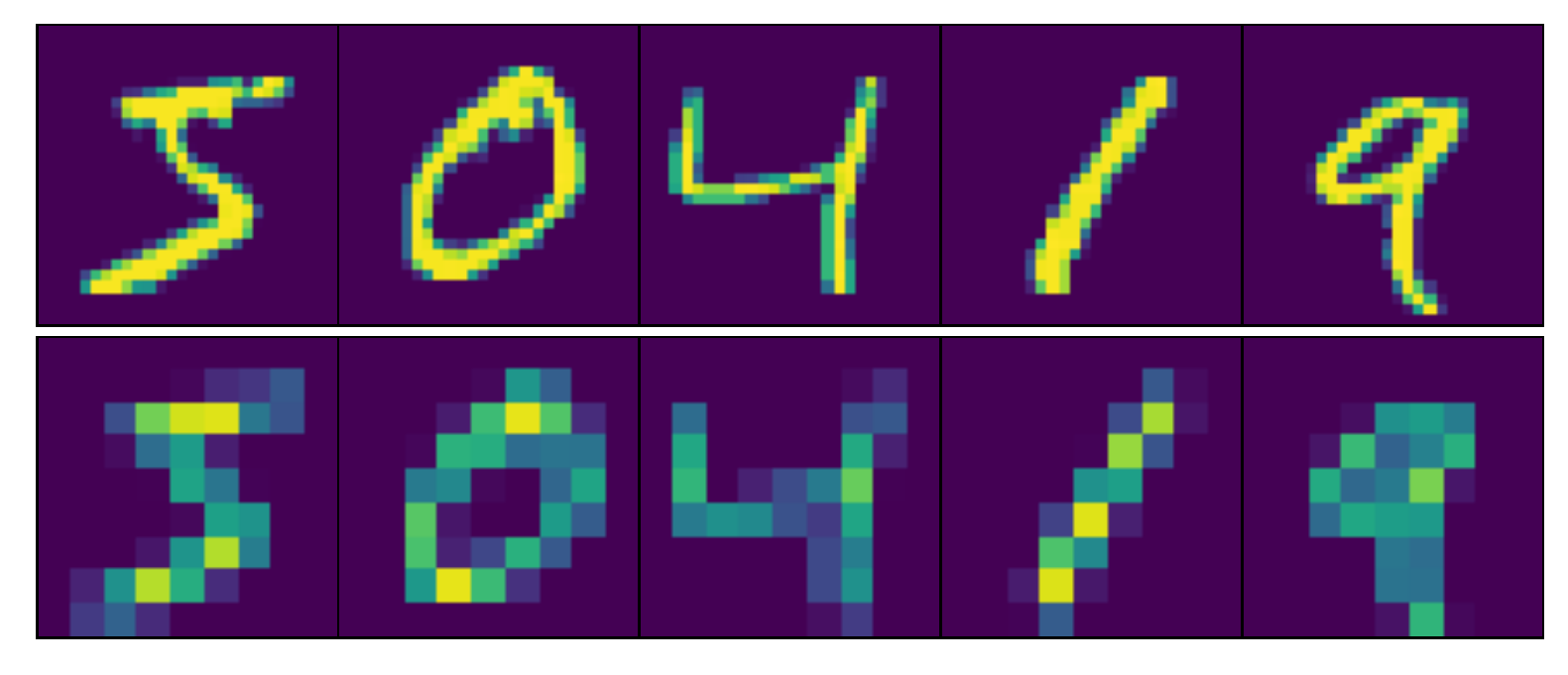} \includegraphics[width=0.48\textwidth]{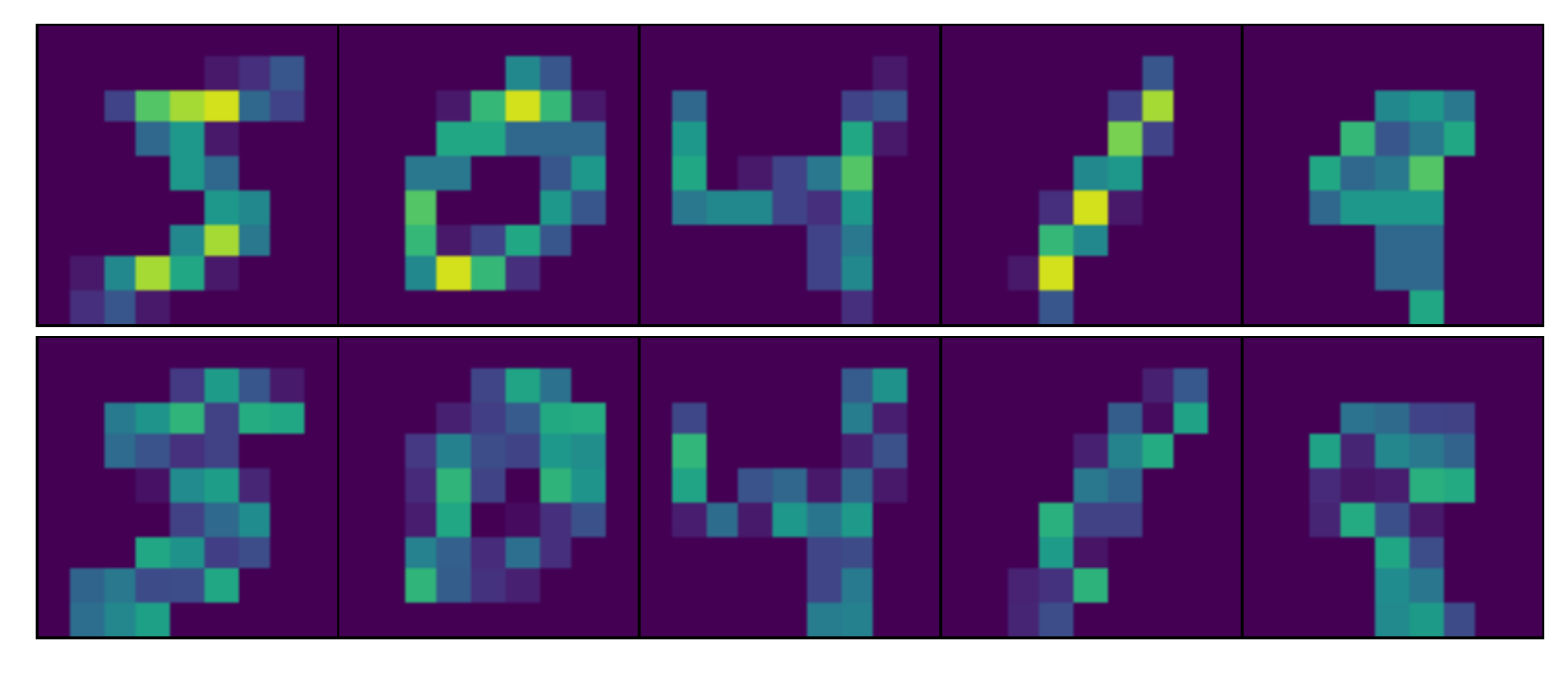}
        \caption{Visualization of the down-scaled and converted images. 
        The top left row shows the first five images of the MNIST training data set.
        The bottom left row shows down-scaled images using $3\times3$ average pooling. 
        The top right row represents the conversion to spikes and back to analog values.
        The bottom right row shows differences between down-scaled and converted images scaled up by a factor of 10.}
        \label{fig:mnist_example}
    \end{figure}
    \begin{figure}[b]
        \centering
        \includegraphics[]{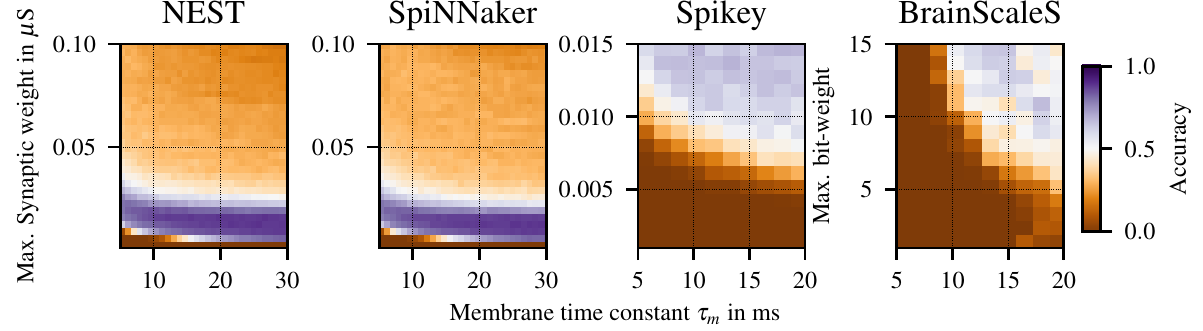}
        \caption{Sweep over the maximal input frequency. Weights for BrainScaleS are set via low level digital weights (0 to 15).}
        \label{fig:sweep_2dim}
    \end{figure}    
    \begin{figure}[t]
        \centering
        \includegraphics[]{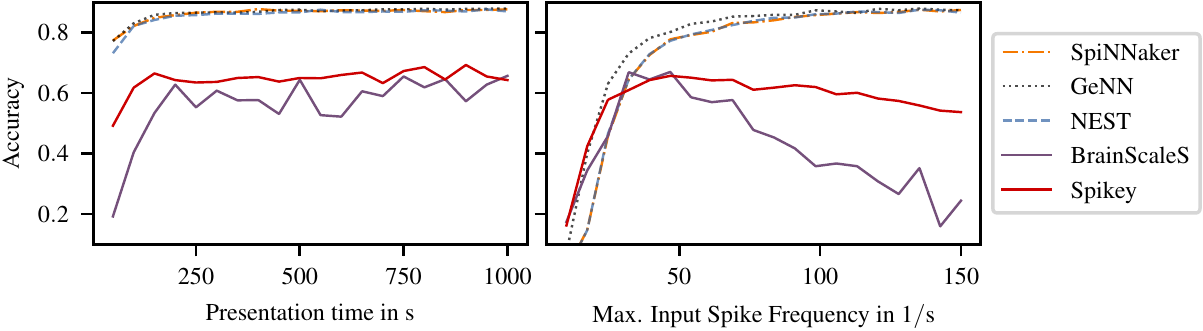}
        \caption{Sweep over the sample presentation time (left) and the maximal input frequency (right)}
        \label{fig:sweep_onedim}
    \end{figure}

    The first two parts of this section present the parameter tuning process used for the converted SNNs.
    Details of four different networks are shown, the smallest one was defined by the restrictions of the Spikey platform, while the remaining networks were picked from the neural architecture search.
    The final part gathers the results for all networks including one model taken from literature.

\subsection{The Spikey Network and Parameter Optimization}
    
    This is the simplest network used in this work.
    As described above, it is motivated by the hardware restriction of the Spikey neuromorphic hardware system and uses a $89\times 100 \times 10$ layout which requires images to be scaled down using $ 3\times 3$ average pooling (cf. Fig. \ref{fig:mnist_example}). 
    These restrictions limit the test-accuracy of the pre-trained network to only $90.13\%$. This serves as the baseline for the following optimizations of the most relevant SNN parameters.
    \begin{itemize}
        \item The \textbf{maximal weight} determines the incoming activity per neuron. 
            If chosen too high, the neuron operates in its non-linear and saturating range near the maximum output frequency.
        \item The \textbf{leakage/membrane time constant} describes the time window in which the neuron integrates incoming input.
            Too small values would require high frequencies for encoding analog values while higher numbers lead to saturation effects.
        \item The \textbf{sample presentation time} increases accuracy with higher values, which in turn require more energy and time.
        \item A higher \textbf{frequency range of input pixels} improves the pixel approximation accuracy, but is subject to saturation of neurons.
    \end{itemize}
    Figure \ref{fig:sweep_2dim} shows parameter sweeps over the two most essential neuron parameters for the training set. 
    The images show large areas of high relative accuracy for the analog platforms.
    On the simulated platforms, one can see the discussed effects of saturating neurons at high weights/time constants. 
    Here, the area of high relative accuracy is rather narrow.
    Therefore, careful parameter tuning has to be done.\looseness=-1
    
    Taking a look at the most relevant conversion parameters, Figure \ref{fig:sweep_onedim} shows the accuracy in relation to the sample presentation time and the maximal spike input frequency.
    First, simulating more than \SI{200}{\milli\second} will result in minor improvements only.
    Analog platforms converge a bit slower (which is partially caused by different neuron parameters used in the simulation), and the benefits of using presentation times larger than \SI{200}{\milli\second} are minor again.
    However, prolonged presentation times can cancel out some of the temporal noise on membrane voltages and synapses.
    Second, all platforms gain significantly from frequencies larger than \SI{40}{\hertz}. 
    However, due to communication constraints in the accelerated analog platforms, the accuracy decreases for values above \SI{60}{\hertz}.
    Here, two bandwidth restrictions may play a major role:
    input spikes are inserted into the digital network using FPGAs.
    Any spike loss is usually reported by the respective software layer.
    However, on the wafer, there might be additional loss in the internal network, which is not reported.
    Output rates of hidden and ouput layers are a second source of potential spike loss which is only partially reported for the Spikey system (by monitoring spike buffers), but happens silently on the BrainScaleS system.
    The Spikey system reports full buffers for larger frequencies, which is why we assume that this is the major cause for spike loss on both systems.\looseness=-1
    
    To reach a high efficiency on larger systems, it is crucial to fully utilize them.
    Therefore, we used several parallel instances of the same network each classifying a separate portion of the data.
    In our setup this is controlled by choosing the batch size: a smaller batch size leads to more independent batches processed in parallel and thus effectively reduces processing time and energy per inference.
    This also avoids idle cores contributing to the energy calculation.
    These system-specific variations in batch size have negligible effects on the classification accuracy. 
    On SpiNNaker, the hardware size and the required number of processor cores per network instance determine the parallelism.
    On GeNN, the working memory required to compile the GPU code is the determining factor.
    The latter is a limitation caused by using separate populations per layer, which could be merged to possibly lead to an increased parallelism of the networks, but not necessarily to increased efficiency.
    Only the Spikey system executes batches sequentially to avoid full spike buffers.

\subsection{NAS Optimized Networks}
    \begin{figure}[t]
        \centering
        \includegraphics[width=\textwidth]{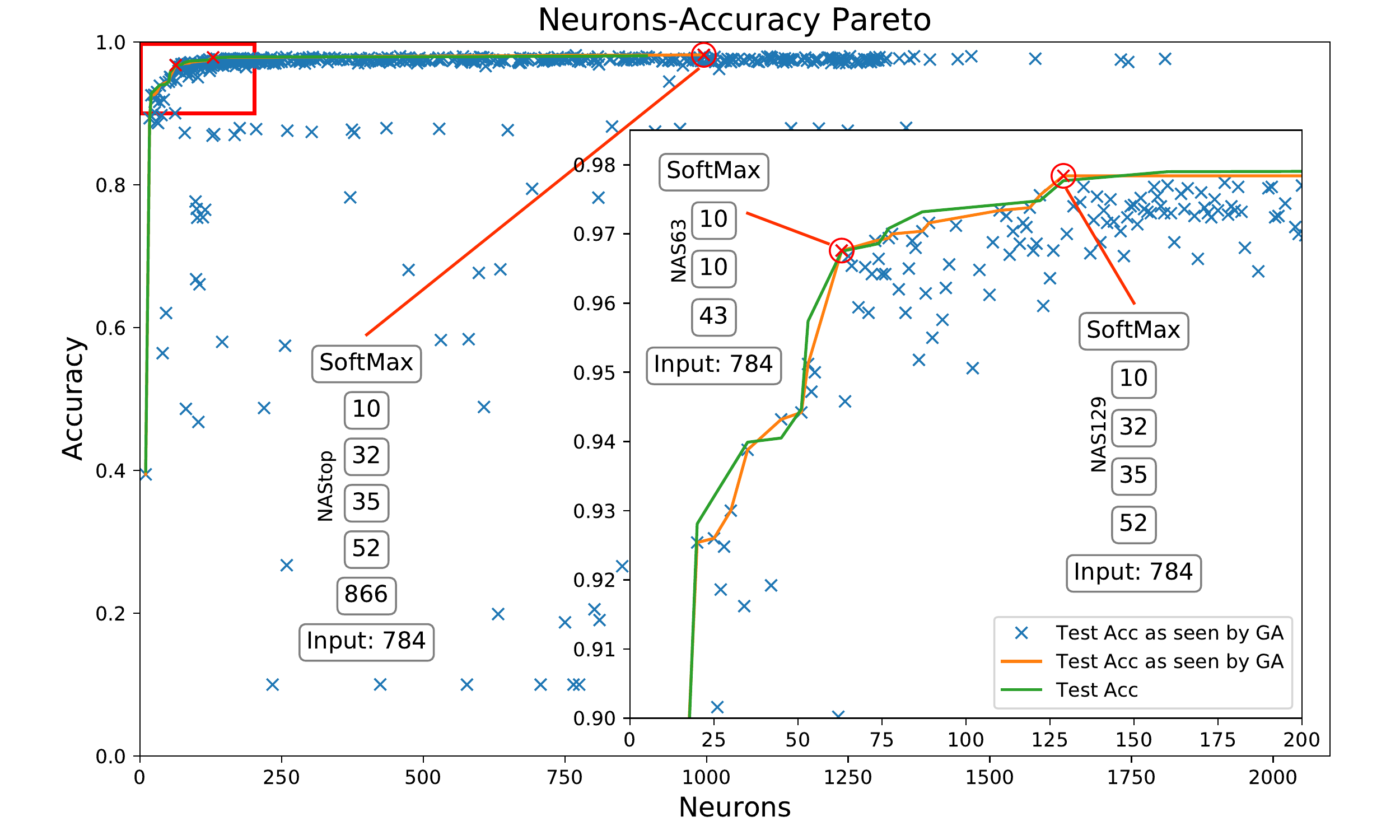}
        \caption{Results of the optimization process. Highlighted are three candidates networks at the pareto front with their respective network layout.}
        \label{fig:nas_results}
    \end{figure}
    
    The optimization process was driven by two major goals: 
    to reach an accuracy larger than 97\% and at the same time to reduce the network size in terms of the number of neurons.
    Results in Figure \ref{fig:nas_results} reveal, that this not necessarily leads to networks with a single hidden layer. 
    Furthermore, the sequential neural networks outperformed all evaluated non-sequential architectures. 
    We have chosen three candidates on the pareto-front for evaluation on neuromorphic hardware:
    \begin{itemize}
        \item the network with the highest evaluation accuracy (\textit{NAStop}, 97.71\%)
        \item the optimal network with the best trade-off (\textit{NAS129}, 97.53\%)
        \item a small network with still sufficient accuracy (\textit{NAS63}, 96,76\%)
    \end{itemize}{}

 \subsection{Benchmark results}
    \begin{table}[p]
    \setlength{\tabcolsep}{0.5em}
    \centering
    \caption{Results from all converted networks.
    Highlighted are the best values per converted network.
    $^\dagger$ Reduced number of neurons per core from its default 255 to 200 and $^\times$ further reduced to 180 together with a slowed-down simulation (factor 2).}
    \begin{tabular}{@{}lrrrrr@{}}
    \toprule
    
    Platform & \begin{tabular}{{@{}c@{}}}\colorbox{white}{Accuracy}\end{tabular} &
    \begin{tabular}{{@{}r@{}}}\colorbox{white}{Conversion}\\\colorbox{white}{Loss}\end{tabular} &
    \begin{tabular}{{@{}r@{}}}\colorbox{white}{Wall clock}\\\colorbox{white}{time}\end{tabular} &
    \begin{tabular}{{@{}r@{}}}\colorbox{white}{Energy per}\\\colorbox{white}{Inference}\end{tabular} & Batchsize \\
    
    & \colorbox{white}{in \%}    & \colorbox{white}{in \%}       & \colorbox{white}{in \SI{}{\milli\second}} & \colorbox{white}{in \SI{}{\milli\joule}} &  \\\midrule

    \multicolumn{6}{c}{\textit{Spikey Network} (ANN accuracy: 90.13\%)}\\\midrule
    Spikey          & \colorbox{white}{65.33}     &\colorbox{white}{24.80} & \colorbox{moonstone}{350} & \colorbox{moonstone}{0.21} & 2500   \\
    Spikey HIL      & \colorbox{white}{84.99}     &\colorbox{white}{5.14} & \colorbox{moonstone}{350} & \colorbox{moonstone}{0.21} & 100    \\
    BrainScaleS     & \colorbox{white}{61.65}     &\colorbox{white}{28.43} & \colorbox{white}{900}     & \colorbox{white}{0.33} & 10000      \\
    BrainScaleS HIL & \colorbox{white}{83.87}     &\colorbox{white}{6.56} & \colorbox{white}{900}     & \colorbox{white}{0.36} & 10000      \\
    SpiNN3          & \colorbox{white}{88.41}     &\colorbox{white}{1.72} & \colorbox{white}{264000}  & \colorbox{white}{79} & 480          \\
    SpiNN5          & \colorbox{white}{88.40}     &\colorbox{white}{1.73} & \colorbox{white}{23100}   & \colorbox{white}{61} & 42           \\
    NEST            & \colorbox{white}{88.98}     &\colorbox{white}{1.15} & \colorbox{white}{70542}   & \colorbox{white}{316} & 2500        \\
    GeNN CPU        & \colorbox{moonstone}{89.11} &\colorbox{moonstone}{1.02} & \colorbox{white}{5070}    & \colorbox{white}{10} & 10000        \\
    GeNN GPU        & \colorbox{white}{88.87}     &\colorbox{white}{1.26} & \colorbox{white}{2623}    & \colorbox{white}{21} & 100  \\\midrule
    
    \multicolumn{6}{c}{\textit{NAS63} (ANN accuracy: 96,76\%)}\\\midrule
    SpiNN3       & \colorbox{white}{96.04}      &\colorbox{white}{0.63} & \colorbox{white}{368500}   & \colorbox{white}{109}     & 670       \\
    SpiNN5       & \colorbox{white}{96.04}      &\colorbox{white}{0.63} & \colorbox{white}{30800}    & \colorbox{white}{80}      & 56        \\
    NEST         & \colorbox{moonstone}{96.37}  &\colorbox{moonstone}{0.30} & \colorbox{white}{217252}   & \colorbox{white}{952}     & 10000     \\
    GeNN CPU     & \colorbox{white}{96.29}      &\colorbox{white}{0.38} & \colorbox{moonstone}{16659}& \colorbox{moonstone}{31}  & 10000     \\
    GeNN GPU     & \colorbox{white}{96.32}      &\colorbox{white}{0.35} & \colorbox{white}{17881}    & \colorbox{white}{145}     & 160 \\\midrule
    
    \multicolumn{6}{c}{\textit{NAS129} (ANN accuracy: 97,53\%)}\\\midrule
    SpiNN3       & \colorbox{white}{96.86}     &\colorbox{white}{0.67} & \colorbox{white}{458700}     & \colorbox{white}{138}   & 834       \\
    SpiNN5       & \colorbox{white}{97.25}     &\colorbox{white}{0.28} & \colorbox{white}{38500}      & \colorbox{white}{105}   & 70        \\
    NEST         & \colorbox{white}{97.10}     &\colorbox{white}{0.43} & \colorbox{white}{263134}     & \colorbox{white}{1247}  & 10000     \\
    GeNN CPU     & \colorbox{moonstone}{97.42} &\colorbox{moonstone}{0.11} & \colorbox{white}{20436}      & \colorbox{moonstone}{38}& 10000     \\
    GeNN GPU     & \colorbox{white}{97.34}     &\colorbox{white}{0.19} & \colorbox{moonstone}{18495}  & \colorbox{white}{153}   & 200 \\\midrule
    
    \multicolumn{6}{c}{ \textit{NAStop}  (ANN accuracy: 97,71\%)}\\\midrule
    SpiNN3$^\dagger$   & \colorbox{white}{96.80}    &\colorbox{white}{0.91} & \colorbox{white}{918500}    & \colorbox{white}{353}     & 1670      \\
    SpiNN5$^\dagger$   & \colorbox{white}{97.42}    &\colorbox{white}{0.29} & \colorbox{white}{82500}     & \colorbox{white}{288}     & 150       \\
    NEST               & \colorbox{white}{97.35}    &\colorbox{white}{0.36} & \colorbox{white}{907869}    & \colorbox{white}{4004}    & 10000     \\
    GeNN CPU           & \colorbox{moonstone}{97.53}&\colorbox{moonstone}{0.18} & \colorbox{white}{96324}     & \colorbox{moonstone}{173} & 10000     \\
    GeNN GPU           & \colorbox{white}{97.51}    &\colorbox{white}{0.20} & \colorbox{moonstone}{21355} & \colorbox{white}{196}     & 265 \\\midrule
    
    \multicolumn{6}{c}{Network from \cite{Diehl2015} (ANN accuracy of 98.84\%)}\\\midrule
    SpiNN3$^\times $   & \colorbox{white}{97.83}     &\colorbox{white}{1.01} & \colorbox{white}{2750000}  & \colorbox{white}{1021}    & 2500  \\
    SpiNN5$^\dagger$   & \colorbox{white}{98.77}     &\colorbox{white}{0.07} & \colorbox{white}{104500}   & \colorbox{white}{407}     & 190   \\
    NEST               & \colorbox{white}{98.82}     &\colorbox{white}{0.02} & \colorbox{white}{3061562}  & \colorbox{white}{13869}   & 10000 \\
    GeNN CPU           & \colorbox{moonstone}{98.86} &\colorbox{moonstone}{-0.02} & \colorbox{white}{314049}   & \colorbox{white}{587}     & 10000 \\
    GeNN GPU           & \colorbox{white}{98.85}     &\colorbox{white}{-0.01} & \colorbox{moonstone}{26632}& \colorbox{moonstone}{293} & 280   \\ \bottomrule
    \end{tabular}
    \label{tab:results}
    \end{table}
    
    Table \ref{tab:results} collects the results for all target platforms.
    Most striking is the energy efficiency of the analog platforms, which is two orders of magnitude higher compared to other simulators.
    Furthermore, HIL training recovers most of the conversion losses found for these platforms (despite the four bit weight accuracy).
    Larger networks have not been evaluated either due to size restrictions, or because combined spike rates of input pixels are too high to get any reasonable results.
    The SpiNNaker system, in both variants, performs on the same efficiency level as a CPU/GPU implementations although its technology is much older (\SI{130}{\nano\meter} vs. \SI{22}{\nano\meter} CPU  vs. \SI{16}{\nano\meter} GPU).
    Furthermore, there is less than one percent loss in accuracy due to the conversion in almost all cases.
    However, for the large networks the system was performing at its limits, and we had to reduce the maximal number of neurons per core. 
    Of course, this can be mitigated by further reducing the number of neurons per core or slowing down the system with respective negative impacts on the energy per inference.
    Interesting differences have been found for NEST: in some cases the accuracy is a bit lower, but the energy per inference is one order higher than for the GeNN CPU simulation.
    The latter is mainly due to the more accurate integrator employed by the NEST simulator (especially the adaptive time step in the integrator), which is also responsible for the significant energy gap between the two CPU simulators NEST and GeNN. 
    Furthermore, the multi-threaded execution of NEST does not reduce the computation time compared to GeNN.
    With the increase of network complexity there is next to no increase in GPU execution time, indicating that despite parallelization of the networks, the GPU is still not utilized fully for the smaller networks (there are 3969-86,760 simultaneously simulated neurons for the GPU depending on the network).
    Still, for the larger networks, the GPU implementation is the fastest simulation available. 
    
    The last network in Table \ref{tab:results} is taken from \cite{Diehl2015}, as the network weights are published within the respective repository.
    The layout is $784\times 1200 \times 1200 \times 10$, and thus it is significantly larger.
    The results show that the SpiNN3 system still operates at its limits (as reported by the software stack) despite the used slow-down.
    The other platforms show nearly the same accuracy with next to no loss in the conversion process.
    Concerning the energy per inference, the larger SpiNNaker platform is slightly better than the CPU implementation, with the GPU being the most efficient platform.

\section{Conclusion and Outlook}
We have demonstrated the capability of all target platforms to simulate converted deep neural networks.
The loss in the conversion process is negligible in many cases, and for analog platforms  Spikey and BrainScaleS we successfully employed retraining to reach high accuracy.
Furthermore, we calculated the used energy-per-inference, quantifying the efficiency vs. accuracy trade-off of analog platforms.
The digital SpiNNaker platform is highly efficient if fully utilized despite the rather old chip manufacturing process, demonstrating the suitability for efficient large-scale simulations.
If primarily simulation time at highest accuracy for not too large networks needs to be optimized, GeNN's GPU backend allow fast and efficient simulation of SNNs.
The approach used in this work is not the most efficient way of using spiking neural networks. 
However, the rate-coding applied here can be replaced with a more efficient time-to-first-spike (TTFS) encoding, using only a few spikes with much faster response times, which has recently been demonstrated on analog hardware \cite{Goltz2019}.
Therefore, the results from this work must be seen as a conservative measure for the relative efficiency of SNNs on neuromorphic hardware.
Furthermore, we did not make use of convolutional networks, because these currently cannot be mapped well to neuromorphic hardware.
For the future of our benchmark suite we plan to include both: networks using TTFS encoding and convolutions.
This will allow us to test more challenging data-sets with larger and more complex networks.\looseness=-1 

\subsubsection*{Funding/Acknowledgment}
The research leading to these results has received funding from the European Union Seventh Framework Programme (FP7) under grant agreement no 604102 and the EU's Horizon 2020 research and innovation programme under grant agreements No 720270 and 785907 (Human Brain Project, HBP).
It has been further supported by the Cluster of Excellence Cognitive Interaction Technology \enquote{CITEC} (EXC 277) at Bielefeld University, which is funded by the German Research Foundation (DFG).
Furthermore, we thank the Electronic Vision(s) group from Heidelberg University and Advanced Processor Technologies Research Group from Manchester University for access to their hardware systems and continuous support and James Knight from the University of Sussex for support regarding our GeNN implementation.
\bibliographystyle{splncs04}
\bibliography{bibexport}

\begin{thebibliography}{10}
\providecommand{\url}[1]{\texttt{#1}}
\providecommand{\urlprefix}{URL }
\providecommand{\doi}[1]{https://doi.org/#1}

\bibitem{Akopyan2015}
Akopyan, F., et~al.: {TrueNorth: Design and Tool Flow of a 65 mW 1 Million
  Neuron Programmable Neurosynaptic Chip}. IEEE Transactions on Computer-Aided
  Design of Integrated Circuits and Systems  \textbf{34}(10),  1537--1557 (10
  2015). \doi{10.1109/TCAD.2015.2474396}

\bibitem{VanAlbada2018}
van Albada, et~al.: {Performance Comparison of the Digital Neuromorphic
  Hardware SpiNNaker and the Neural Network Simulation Software NEST for a
  Full-Scale Cortical Microcircuit Model}. Frontiers in Neuroscience
  \textbf{12} (5 2018). \doi{10.3389/fnins.2018.00291}

\bibitem{Cao2015}
Cao, Y., et~al.: {Spiking Deep Convolutional Neural Networks for
  Energy-Efficient Object Recognition}. International Journal of Computer
  Vision  \textbf{113}(1),  54--66 (5 2015). \doi{10.1007/s11263-014-0788-3}

\bibitem{davies2019benchmarks}
Davies, M.: {Benchmarks for progress in neuromorphic computing}. Nature Machine
  Intelligence  \textbf{1}(9),  386--388 (2019).
  \doi{10.1038/s42256-019-0097-1}

\bibitem{davies2018loihi}
Davies, M., et~al.: {Loihi: A Neuromorphic Manycore Processor with On-Chip
  Learning}. IEEE Micro  \textbf{38}(1),  82--99 (2018).
  \doi{10.1109/MM.2018.112130359}

\bibitem{Davison2008}
Davison, A.P.: {PyNN: a common interface for neuronal network simulators}.
  Frontiers in Neuroinformatics  \textbf{2}(January), ~11 (2008).
  \doi{10.3389/neuro.11.011.2008}

\bibitem{Diehl2015}
Diehll, P.U., et~al.: {Fast-Classifying, High-Accuracy Spiking Deep Networks
  Through Weight and Threshold Balancing}. Proceedings of the International
  Joint Conference on Neural Networks  \textbf{2015-Septe} (2015).
  \doi{10.1109/IJCNN.2015.7280696}

\bibitem{Furber2013}
Furber, S.B., et~al.: {Overview of the SpiNNaker system architecture}. IEEE
  Transactions on Computers  \textbf{62}(12),  2454--2467 (12 2013).
  \doi{10.1109/TC.2012.142}

\bibitem{gewaltig2007nest}
Gewaltig, M.O., Diesmann, M.: {{\{}NEST{\}} ({\{}NEural{\}} Simulation Tool)}.
  Scholarpedia  \textbf{2}(4), ~1430 (2007)

\bibitem{Goltz2019}
G{\"{o}}ltz, J., et~al.: {Fast and deep neuromorphic learning with
  time-to-first-spike coding}  (12 2019). \doi{10.1145/3381755.3381770}

\bibitem{Homburg2019}
Homburg, J.D., et~al.: {Constraint Exploration of Convolutional Network
  Architectures with Neuroevolution}. In: Advances in Computational
  Intelligence. pp. 735--746. Springer International Publishing (2019).
  \doi{10.1007/978-3-030-20518-8{\_}61}

\bibitem{Jordan2019}
Jordan, J., et~al.: {NEST 2.18.0}  (6 2019). \doi{10.5281/ZENODO.2605422}

\bibitem{Knight2018}
Knight, J.C., Nowotny, T.: {GPUs Outperform Current HPC and Neuromorphic
  Solutions in Terms of Speed and Energy When Simulating a Highly-Connected
  Cortical Model}. Frontiers in Neuroscience  \textbf{12}(December),  1--19
  (2018). \doi{10.3389/fnins.2018.00941}

\bibitem{Maass1997}
Maass, W.: {Networks of spiking neurons: The third generation of neural network
  models}. Neural Networks  \textbf{10}(9),  1659--1671 (12 1997).
  \doi{10.1016/S0893-6080(97)00011-7}

\bibitem{moradi2018scalable}
Moradi, S., et~al.: {A scalable multicore architecture with heterogeneous
  memory structures for Dynamic Neuromorphic Asynchronous Processors (DYNAPs)}.
  IEEE transactions on biomedical circuits and systems  \textbf{12}(1),
  106--122 (2018). \doi{10.1109/TBCAS.2017.2759700}

\bibitem{Neckar2019}
Neckar, A., et~al.: {Braindrop: A Mixed-Signal Neuromorphic Architecture With a
  Dynamical Systems-Based Programming Model}. Proceedings of the IEEE
  \textbf{107}(1),  144--164 (1 2019). \doi{10.1109/JPROC.2018.2881432}

\bibitem{2941207}
Ostrau, C., et~al.: {Comparing Neuromorphic Systems by Solving Sudoku
  Problems}. In: Conference Proceedings: 2019 International Conference on High
  Performance Computing {\&} Simulation (HPCS). IEEE (2019).
  \doi{10.1109/HPCS48598.2019.9188207}

\bibitem{2941831}
Ostrau, C., et~al.: {Benchmarking of Neuromorphic Hardware Systems}. In:
  Proceedings of the Neuro-Inspired Computational Elements Workshop.
  Association for Computing Machinery (ACM) (2020).
  \doi{10.1145/3381755.3381772}

\bibitem{Petrovici2014a}
Petrovici, M.A., et~al.: {Characterization and compensation of network-level
  anomalies in mixed-signal neuromorphic modeling platforms}. PLoS ONE
  \textbf{9}(10) (2014). \doi{10.1371/journal.pone.0108590}

\bibitem{pfeil2013six}
Pfeil, T., et~al.: {Six networks on a universal neuromorphic computing
  substrate}. Frontiers in Neuroscience  \textbf{7}(7 FEB), ~11 (2013).
  \doi{10.3389/fnins.2013.00011}

\bibitem{Rhodes2020}
Rhodes, O., et~al.: {Real-time cortical simulation on neuromorphic hardware}.
  Philosophical Transactions of the Royal Society A: Mathematical, Physical and
  Engineering Sciences  \textbf{378}(2164),  20190160 (2 2020).
  \doi{10.1098/rsta.2019.0160}

\bibitem{schemmel2010waferscale}
Schemmel, J., et~al.: {A wafer-scale neuromorphic hardware system for
  large-scale neural modeling}. In: Proceedings of 2010 IEEE International
  Symposium on Circuits and Systems. pp. 1947--1950 (2010).
  \doi{10.1109/ISCAS.2010.5536970}

\bibitem{schmitt2017neuromorphic}
Schmitt, S., et~al.: {Neuromorphic hardware in the loop: Training a deep
  spiking network on the BrainScaleS wafer-scale system}. In: 2017
  International Joint Conference on Neural Networks (IJCNN). pp. 2227--2234.
  IEEE (5 2017). \doi{10.1109/IJCNN.2017.7966125}

\bibitem{stockel2017binary}
St{\"{o}}ckel, A., et~al.: {Binary associative memories as a benchmark for
  spiking neuromorphic hardware}. Frontiers in computational neuroscience
  \textbf{11}(August), ~71 (2017). \doi{10.3389/fncom.2017.00071}

\bibitem{Yavuz2016}
Yavuz, E., et~al.: {GeNN: a code generation framework for accelerated brain
  simulations.} Scientific reports  \textbf{6}(June 2015),  18854 (2016).
  \doi{10.1038/srep18854}

\end{thebibliography}

\end{document}